\documentclass{article}

\PassOptionsToPackage{numbers}{natbib}
    \usepackage[final]{neurips_2019}

\usepackage[utf8]{inputenc} % allow utf-8 input
\usepackage[T1]{fontenc}    % use 8-bit T1 fonts
\usepackage{hyperref}       % hyperlinks
\usepackage{url}            % simple URL typesetting
\usepackage{booktabs}       % professional-quality tables
\usepackage{amsfonts}       % blackboard math symbols
\usepackage{nicefrac}       % compact symbols for 1/2, etc.
\usepackage{microtype}      % microtypography

\usepackage[title]{appendix}
\usepackage{amsmath,amssymb,amsfonts}

\DeclareMathOperator*{\argmin}{arg\,min}
\newcommand*{\argminl}{\argmin\limits}

\usepackage{algorithmic}
\usepackage{algorithm}
\usepackage{graphicx}
\usepackage{textcomp}
\usepackage{xcolor}
\usepackage{fancyhdr}

\title{Bottom-Up Meta-Policy Search}

\author{%
  Luckeciano C. Melo \\
  Autonomous Computational Systems Lab\\
  Computer Science Division\\
  Aeronautics Institute of Technology\\
  São José dos Campos, Brazil \\
  \texttt{luckeciano@gmail.com} \\
   \And
   Marcos R. O. A. Maximo \\
  Autonomous Computational Systems Lab\\
  Computer Science Division\\
  Aeronautics Institute of Technology\\
  São José dos Campos, Brazil \\
   \texttt{mmaximo@ita.br} \\
   \And
   Adilson M. da Cunha \\
   Software Engineering Research Group \\
   Computer Science Division \\
   Aeronautics Institute of Technology \\
   São José dos Campos, Brazil \\
   \texttt{cunha@ita.br} \\
}

\begin{document}

\maketitle

\begin{abstract}
  Despite of the recent progress in agents that learn through interaction, there are several challenges in terms of sample efficiency and generalization across unseen behaviors during training. To mitigate these problems, we propose and apply a first-order Meta-Learning algorithm called Bottom-Up Meta-Policy Search (BUMPS), which works with two-phase optimization procedure: firstly, in a meta-training phase, it distills few expert policies to create a meta-policy capable of generalizing knowledge to unseen tasks during training; secondly, it applies a fast adaptation strategy named Policy Filtering, which evaluates few policies sampled from the meta-policy distribution and selects which best solves the task. We conducted all experiments in the RoboCup 3D Soccer Simulation domain, in the context of kick motion learning. We show that, given our experimental setup, BUMPS works in scenarios where simple multi-task Reinforcement Learning does not. Finally, we performed experiments in a way to evaluate each component of the algorithm.
\end{abstract}

\section{Introduction}

In recent years, Machine Learning have been able to achieve or even surpass the human performance in several challenges regarding machine perception, planning and reasoning, control, and multi-agent strategy \cite{DBLP:journals/corr/LuT14, DBLP:journals/corr/XiongDHSSSYZ16a, openaifive, alphastarblog}. It also achieved surprisingly results in controlling agent locomotion in simulation \cite{DBLP:journals/corr/HeessTSLMWTEWER17}, or in real world \cite{xie2019iterative}. In this way, the learning field of combining techniques from Deep Learning and Reinforcement Learning (RL) appears as a great candidate in the search of General Artificial Intelligence.

Nevertheless, beyond the success of these techniques, there are several challenges to improve them in terms of sample complexity and generalization, specially in robotic control, where we lead with hardware and/or physical restrictions in order to learn robust policies for a diversity of behaviors. In this context, simulated environments are very useful to consistently evaluate new algorithms for that purpose.

In the light of these ideas, this work contributes by proposing an algorithm based on Imitation Learning and Meta-Learning to learn and optimize policies for humanoid robot control, being initially evaluated in the context of robot soccer. The purpose of Bottom-Up Meta-Policy Search (BUMPS) is to generalize the knowledge from few expert policies to learn how to learn a diversity of similar control behaviors.

The remaining of this work is organized as follows. Section \ref{sec:relatedwork} presents related work. Section \ref{sec:background} provides theoretical background. In Section \ref{sec:bumps} we present BUMPS in a theoretical perspective. Furthermore, Section \ref{sec:results_and_discussion} presents simulation results to validate our approach. Finally, Section \ref{sec:conclusion} concludes and shares our ideas for future work.

\section{Related Work}\label{sec:relatedwork}

In terms of Imitation Learning -- a technique where the agent learns a policy by acquiring skills from observing demonstrations \cite{DBLP:journals/corr/DuanASHSSAZ17} -- there are great successful stories about its applications in complex domains: \citet{Abbeel:2010:AHA:1894938.1894944} elaborated algorithms for learning trajectory-based task specification from demonstrations and for modeling of helicopter's dynamics, in order to design autonomous flight controllers; \citet{5152577} applied this technique to make a robot learn how to play tennis; and \citet{learnarms} made an agent performs human-like reaching motions. In the context of robot soccer, \citet{DBLP:journals/corr/abs-1901-00270} applied Behavior Cloning to learn keyframe kick motion in a neural network, and \citet{ICML2019-pavse} proposed a method of combining reinforcement learning and imitation from observation to perform imitation using a single expert demonstration. Survey articles about Imitation Learning include \cite{Schaal99isimitation, Billard2008, Argall:2009:SRL:1523530.1524008}.
 
 In terms of Meta-Learning -- the idea of learn how to learn efficiently \cite{schmidhuber:1987:srl, bengiometalearning, Thrun1998} -- there are several studies that conceived this approach in distinct ways: by investigating automatic tuning of hyperparameters \cite{PMID:12371519}; by searching a meta-learner representation that uses a specific model architecture, using recurrent policies \cite{metalearning, DBLP:journals/corr/WangKTSLMBKB16} or attention mechanism and causal convolutions \cite{snail}; lastly, by learning a meta-gradient via optimization and then quickly adapting the policy with few gradient steps \cite{DBLP:conf/iclr/RaviL17, DBLP:journals/corr/FinnAL17}. Other recent ideas of Meta-Learning, specifically to Reinforcement Learning context, is to use a learned critic network that provides gradients to the policy \cite{DBLP:journals/corr/SungZXHY17} or using a planner and an adaptable model for model-based RL \cite{DBLP:journals/corr/abs-1802-04821}.
 
 In terms of meta-policy search, \citet{DBLP:journals/corr/abs-1904-00956} also proposed an algorithm that use similar ideas of leveraging supervised imitation learning rather than relying on high-variance algorithms such as policy gradients, decoupling the problem of obtaining expert trajectories for every task from the problem of learning a fast adaptation algorithm. In contrast, our approach relies on a contextual policy to learn a rich representation of tasks and then be able to quickly adapt to unseen ones by using a meta-gradient between contexts, instead of minimizing a meta-objective.
 
 Our approach is also related to few-shot imitation learning \cite{DBLP:journals/corr/DuanASHSSAZ17}, where supervised learning is used for meta-optimization. However, in contrast to such method, our work learn using only reward signals and does not require demonstrations for unseen tasks. Finally, our approach is also related to multi-task RL, specifically with works where there is a training procedure for each training task to provide expert policies \cite{Levine:2016:ETD:2946645.2946684, DBLP:journals/corr/abs-1711-09874}. However, we use those policies to train a meta-learner that not just generalizes to unseen tasks, but also could be used to quickly adapt for new ones.
 
 \section{Preliminaries}\label{sec:background}
 
 \subsection{Markov Decision Processes}
We address policy learning in continuous state and action spaces. We consider the problem of learning a control policy as a Markov Decision Process (MDP), defined by the tuple $M = (\mathcal{S}, \mathcal{A}, \mathcal{P}, r, \rho_{0}, \gamma, T)$, in which $\mathcal{S}$ is a state space, $\mathcal{A}$ is an action space, $\mathcal{P}: \mathcal{S} \times \mathcal{A} \times \mathcal{S} \rightarrow \mathcal{R}_{+}$ a transition probability distribution, $r: \mathcal{S} \times \mathcal{A} \rightarrow [-r_{bound}, +r_{bound}]$ a bounded reward function, $\rho_{0} : \mathcal{S} \rightarrow \mathcal{R}_{+}$ an initial state distribution, $\gamma \in [0, 1]$ a discount factor and $T$ the length of the finite horizon.

During policy optimization, we typically optimize a policy $\pi_{\boldsymbol{\theta}} : \mathcal{S} \times \mathcal{A} \rightarrow \mathcal{R}_{+}$, parameterized by $\boldsymbol{\theta}$, with the objective of maximizing the cumulative reward throughout the episode:

\begin{equation}
\max_{\boldsymbol{\theta}} \mathbb{E}_{\tau}\Big[\sum_{t=0}^{T} \gamma^{t} r(s_{t}, a_{t})\Big],
\end{equation}
where $\tau$ denotes the trajectory, $s_{0} \sim \rho_{0}(s_{0})$, $a_{t} \sim \pi_{\boldsymbol{\theta}}(a_{t} \mid s_{t})$, and $s_{t+1} \sim \mathcal{P}(s_{t+1} \mid s_{t}, a_{t})$. 

\subsection{Supervised Learning for Imitation}

In the context of supervised imitation learning (Behavior Cloning), there is an expert policy $\pi(s \mid a)$ for a given MDP $M$. Our purpose is to mimic this policy, i.e, learn its actions given states. To achieve this, we collect roll-outs from this expert policy to create a dataset $\mathcal{D}$ composed by tuples $(s, a)$ of states and actions. We then parameterize a new policy to learn by minimizing the expectation of negative log-likelihood w.r.t expert policy data, as shown in Equation \eqref{eq:bcloss}:

\begin{equation}\label{eq:bcloss}
J(\boldsymbol{\theta}, \mathcal{D})_{BC}  = -\mathbb{E}_{(s_{t}, a_{t}) \sim \mathcal{D}} \log{\pi (a_{t} \mid s_{t})}
\end{equation}

\subsection{Meta-Learning}

In the meta-learning formulation, there is a distribution over tasks $p(\mathcal{T})$, where each task $\mathcal{T}_{i} \sim p(\mathcal{T})$ is defined by inputs $s_{t}$, outputs $a_{t}$, a loss function $\mathcal{L}_{i}(s_{t}, a_{t})$ and an episode length $H_{i}$. The meta-learning model, namely meta-learner, with parameters $\boldsymbol{\theta}$, has the objective to minimize the expected loss of distribution $p(\mathcal{T})$ w.r.t $\boldsymbol{\theta}$:

\begin{equation}\label{eq:optlossmetalearning}
\min_{\boldsymbol{\theta}}\mathbb{E}_{\mathcal{T}_{i} \sim p(\mathcal{T})} \Big[ \sum_{t = 0}^{H_{i}} \mathcal{L}_{i}(s_{t}, a_{t}) \Big]
\end{equation}

The meta-learner is trained in a set of meta-training tasks $\mathcal{T}_{tr}$, solving the optimization problem from Equation \eqref{eq:optlossmetalearning}. During test phase, a new set of meta-test tasks $\mathcal{T}_{val}$ are sampled to evaluate generalization.

In the RL context, it is possible to formulate meta-learning in terms of Markov Decision Processes, which is closely related to the experiments of this work. Formally, we have a distribution over a set of MDPs $\mathcal{M}$ that we assume known. An agent then optimize a meta-policy $\pi_{\boldsymbol{\theta}}$ by interacting with a set of MDP environments (the tasks), expecting to perform well on average and thus generalize to unseen MDPs sampled from the same distribution $\mathcal{M}$ during testing phase. The objective is averaged across all training MDPs, which reflects the prior that we would like to distill into the agent \cite{metalearning}. It is also valid in the context of POMDPs.

\section{Bottom-Up Meta-Policy Search}\label{sec:bumps}

Bottom-Up Meta-Policy Search (BUMPS) is an algorithm that uses the knowledge of expert policies in single tasks to train a contextual meta-policy that generalizes for unseen tasks. The method is based on the idea of ``Bottom-Up Learning" \cite{Sun2012}, where learning happens firstly in implicit knowledge and then in explicit knowledge (i.e, through ``extracting" implicit knowledge).

The idea of contextual meta-policy comes from the fact we express a policy as $\pi_{\theta}(a_{t} | s_{t}, \omega)$, i.e, we give the task context as input. The meta-training phase then explore the task structure to create a rich representation of shared features based on contextualization. Intuitively, we have the following problem: if we learn tasks whose geometric targets are $a$ and $c$ (such as $a < c$), then we implicitly knows that the representation of policy for target $b$ (such that $a < b < c$) is something approximately ``between" $a$ and $c$.  In this way, BUMPS exploits the geometrical representation of optimal solution manifolds in parameter space. This meta-policy is more robust as it learns MDPs with other contexts, better creating the geometric representation of contextualization.

During meta-training, instead of optimizing a meta-objective such as in MAML \cite{DBLP:journals/corr/FinnAL17}, we use a first-order Behavior Cloning loss (Equation \eqref{eq:bcloss}) within respect to aggregated dataset of several tasks. In this way, we reduce the agent-environment interaction, thus improving sample efficiency. Additionally, we use gradients from supervised learning, which have lower variance and therefore are more stable than RL gradients \cite{DBLP:journals/corr/NorouziBCJSWS16}.

The generalization comes from the fact that such parameters can also solve unseen tasks due to proximity between optimal policies with near contexts. However, we can not guarantee that the best policy for a task will be in the right context, but we can sample some policies in the neighborhood and evaluate them accordingly to a performance metric, and then copying the best policy to a specific context. We call such approach as Policy Filtering. We present Bottom-Up Meta-Policy Search meta-training in Algorithm \ref{alg:bumps} and the Policy Filtering as meta-testing in Algorithm \ref{alg:bumpsmetatest}. 

\begin{algorithm}[H]
	\caption{Bottom-Up Meta-Policy Search Algorithm (Meta-Training)}
	\begin{algorithmic}[1]
		\REQUIRE Distribution over tasks $p({\mathcal{T}})$
		\STATE Sample batch of meta-training tasks
		 $\mathcal{T}_{i} \sim p({\mathcal{T}})$
		\STATE Initialize $\mathcal{D} = \{\}$
		\STATE \textbf{foreach} meta-training task $\mathcal{T}_{i}$ \textbf{do}
		\STATE \hspace{5mm} Train an expert policy $\pi_{\mathcal{T}_{i}}^{*}$ for $\mathcal{T}_{i}$ using RL
		\STATE \hspace{5mm} Sample $K$ trajectories $\mathcal{D}_{\mathcal{T}_{i}} = \{(s_{1}, a_{1}), ...,(s_{H}, a_{H})\}$ using $\pi_{\mathcal{T}_{i}}^{*}$ in $\mathcal{T}_{i}$
		\STATE \hspace{5mm} Contextualize trajectories using $\mathcal{T}_{i}$ goal $\mathcal{D}_{\mathcal{T}_{i}}(\omega) = \{(\mathcal{T}_{i}, s_{1}, a_{1}), ...,(\mathcal{T}_{i},s_{H}, a_{H}) \}$
		\STATE \hspace{5mm} Aggregate $\mathcal{D} \leftarrow \mathcal{D} \bigcup \mathcal{D}_{\mathcal{T}_{i}}(\omega)$
		\STATE \textbf{end for}
		
		\STATE Initialize $\boldsymbol{\theta}$ randomly
		
		\STATE \textbf{while} not done \textbf{do}
		\STATE \hspace{5mm} Sample mini-batch of expert policies dataset $\mathcal{D}^{*} \sim \mathcal{D}$ 
		\STATE \hspace{5mm} Evaluate $\nabla_{\boldsymbol{\theta}} J(\boldsymbol{\theta}, \mathcal{D}^{*})_{BC}$ using Equation \eqref{eq:bcloss}
		\STATE \hspace{5mm} Update parameters with gradient descent $\boldsymbol{\theta} \leftarrow \boldsymbol{\theta} - \alpha \nabla_{\boldsymbol{\theta}} J(\boldsymbol{\theta}, \mathcal{D}^{*})_{BC}$
		\STATE \textbf{end while}
	\end{algorithmic}
	\label{alg:bumps}
\end{algorithm}

\begin{algorithm}[H]
	\caption{BUMPS Policy Filtering (Meta-Testing)}
	\begin{algorithmic}[1]
		\REQUIRE Meta-Policy $\Pi_{\boldsymbol{\theta}}$
		\REQUIRE Meta-Testing task $\mathcal{T}_{test}$
		\REQUIRE Metric loss $\mathcal{L}_{test}$
		\STATE Sample single-task policies $\{\pi_{\boldsymbol{\theta}}^{i} \} = \Pi_{\boldsymbol{\theta}}(\omega)$, where $\omega \sim p(\mathcal{T})$
		\STATE \textbf{foreach} policy $\pi_{\boldsymbol{\theta}}^{i}$ \textbf{do}
		\STATE \hspace{5mm} Evaluate $\pi_{\boldsymbol{\theta}}^{i}$ accordingly to a metric loss $\mathcal{L}_{test}^{i}(\boldsymbol{\theta})$ 
		\STATE \hspace{5mm} $\mathcal{L}_{test}(\boldsymbol{\theta}) \leftarrow \mathcal{L}_{test}(\boldsymbol{\theta}) \bigcup \{ \mathcal{L}_{test}^{i}(\boldsymbol{\theta}) \}$
		\STATE \textbf{end for}
		\STATE Filter $\pi^{*}_{\boldsymbol{\theta}} = \argminl_{\pi_{\boldsymbol{\theta}}} \{\mathcal{L}_{test}(\boldsymbol{\theta})\}$
	\end{algorithmic}
	\label{alg:bumpsmetatest}
\end{algorithm}

\begin{figure}[!htbp]
	\centering
	\includegraphics[ width=0.6\textwidth]{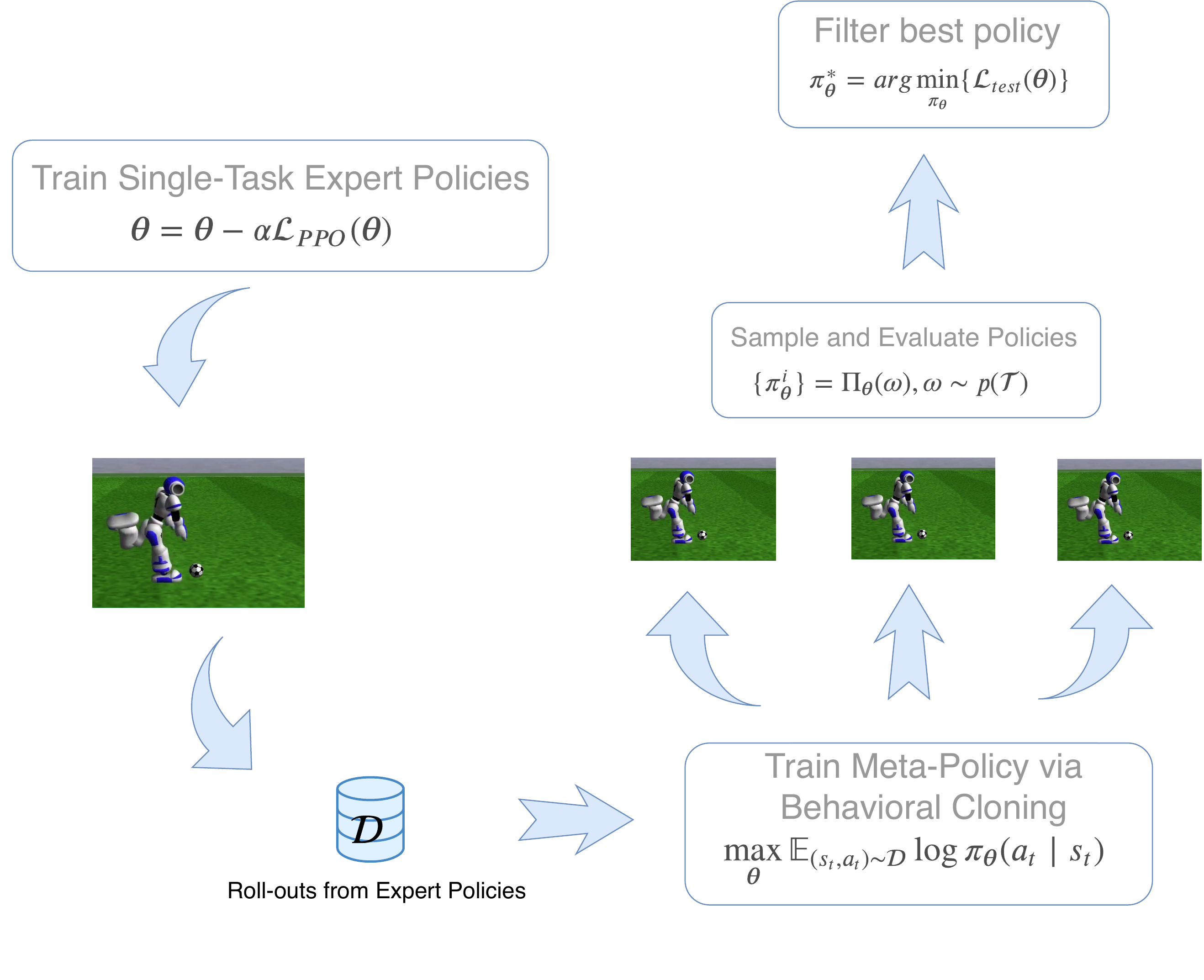}
	\caption{Bottom-Up Meta-Policy Search.}
	\label{fig:bumps}
\end{figure}

Figure \ref{fig:bumps} illustrates how BUMPS works. Before detailing how experiments are conduct in each step of this algorithm, we will explain why BUMPS works.

\subsection{Explaining BUMPS: Meta-Gradient through Contexts}\label{sec:metagradientcontext}

BUMPS optimization has inspirations in Reptile algorithm \cite{DBLP:journals/corr/abs-1803-02999}. The latter conducts few gradient steps towards one task and then compute a kind of  ``meta-gradient" from the parameters before and after such steps. In this way, it learns to use information of higher order derivatives. This process keeps happening across tasks, which leads to a final representation that is close (in Euclidean distance) to the optimal solution manifolds of training tasks.

In contrast, BUMPS trains by using all meta-training tasks roll-outs, using contextualization. This basically solves meta-training tasks using the meta-policy in each meta-training task context. Furthermore, we can assume that a geometric representation of contextualization closely relates to geometric representation of optimal manifolds. Thus, by optimizing meta-training tasks jointly we will also have good representations for meta-testing tasks that are close to them.

Nevertheless, the representation of meta-testing tasks will not necessarily converge to its optimal manifold, then requiring the Policy Filtering approach. Although we assume this close relationship between context and manifold geometry, there is no linearity in the latter. But we then can sample several contexts in the neighborhood and evaluate them to check if they are inside such optimal manifold.

\begin{figure}[!htpb]
	\centering
	\includegraphics[width=0.6\textwidth]{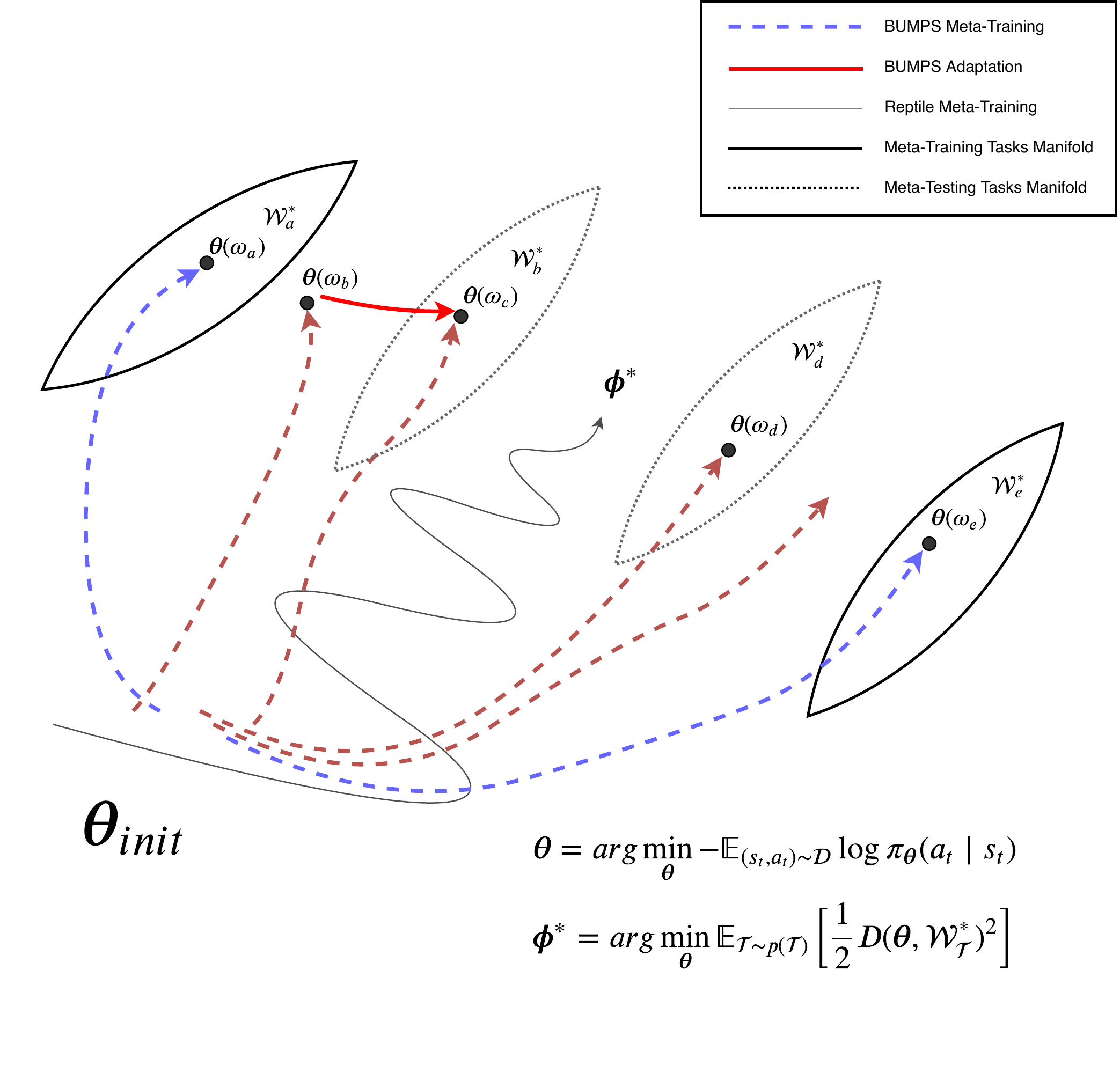}
	\caption{BUMPS and Reptile Meta-Training. In this representation, we consider five tasks $a, b, c, d$ and $e$ and optimal manifolds for $a, b, d$ and $e$. The meta-policy trained by BUMPS and Reptile are parameterized, respectively, by $\boldsymbol{\theta}^{*}$ and $\boldsymbol{\phi}^{*}$. $\boldsymbol{\theta_{init}}$ represents an initial parametrization of policies before training and the notation $\boldsymbol{\theta(\omega_{task})}$ represents the parameterization $\boldsymbol{\theta}$ in the context of a specific task. As we can observe, Reptile training will approximate the parameter set to optimal solution manifolds of meta-training tasks by minimizing the distance between them -- represented by $D(\boldsymbol{\theta}, \mathcal{W}^{*}_{\mathcal{T}})$. However, such parameters will not necessarily converge to these manifolds. On the other hand, BUMPS will conduct their parameters to different regions depending on the context. In the case of meta-training tasks, it will solve them (represented by blue dashed arrows). In the case of meta-testing tasks, however, it will lead to a variety of regions (red dashed arrows): in some cases inside optimal manifold (task $d$), in others only near. To address this problem, we can just sample some contexts and then perform such ``meta-gradient" towards the context inside optimal manifold (as represented by red full arrow).}
    
	\label{fig:bumpsreptile}
\end{figure}

When we found the context that is closer to such manifold (i.e, better solves the task), we then apply the idea of meta-gradient from reptile, but instead using it from parameter sets before and after few training steps, we use between context representations. Figure \ref{fig:bumpsreptile} illustrates and explains BUMPS and a Reptile standard training in parameter set space.

\subsection{Implementation}

In this subsection, we describe our implementation in each step of the BUMPS algorithm. The code is publicly available for the sake of reproducibility.\footnote{\url{https://github.com/luckeciano/bumps}}

\subsubsection{Single-Task Expert Policies}

As first part of the BUMPS algorithm, we need to obtain few single-task expert policies in order to collect roll-outs and conduct meta-training. The algorithm does not impose a specific procedure to this part, so it is possible to use demonstrations or train an expert agent via RL. In our case, since we use a simulated environment, we obtained these policies using RL via Proximal Policy Optimization (PPO) \cite{ppoalgorithm}.

Nevertheless, in some hard tasks, due to the MDP constraints (sparse or delayed reward, high dimensional action space), it is challenging and computationally costly to pure RL techniques obtain policies that achieve high performance. Therefore, in our experiments, we experimented a two-phase optimization procedure, where we first use imitation learning to mimic a initial motion and then optimizes via PPO towards the task. 
% \citet{mscluck} conducted several experimental analyzes in a similar evaluation task used in this work to conclude that this procedure leaded to better performance if compared with pure RL trained policies or the just imitated motion. 

\subsubsection{Training Meta-Policy and Policy Filtering}
During meta-policy training, we firstly collected a single trajectory from each expert policy, in order to show that BUMPS can learn efficiently with few demonstrations. We then added the contextual variable to each trajectory, as an additional input that represented the task. 

The meta-policy parameterization (represented as a neural network), as well as the hyperparameters from optimization are chosen following the same ideas of a standard supervised learning setup. We present these details regarding our experiments in Appendix \ref{ap:hypers}. We also analyzed network depth and ensembles to improve final results. 

During Policy Filtering, we solve a specific task $\mathcal{T}_{test}$ by considering its context $\omega_{\mathcal{T}_{test}}$ as central policy and also sampling other contextual policies from the neighborhood of such context. In our experiments, we also evaluate how the number of samples improves the final performance. Finally, we evaluate all of them accordingly to the simple final target error as online metric, choosing as solution the policy which has the lowest value for it.

% we sampled all tasks in the previously established interval, spaced by 0.1 (i.e, $7.0, 7.1, \dots, 17.9, 18.0$). The evaluation scenario of sampled policies is the same conducted to evaluate single-task expert policies: evaluate kick error mean across one hundred repetitions. As metric loss, we consider the mean absolute error between target and the final distance from kick. 

% We trained during 150k epochs, using initial learning rate of $3 \times 10^{-4}$, decaying by step factor of 0.8 and minimizing mean absolute loss by using Adam \cite{adam2014}. Additional hyperparameters can be found in the released code\footnote{\label{bumps}\url{https://github.com/luckeciano/deep-rl-humanoid-motions-masters/tree/master/BUMPS}}.

% In terms of neural network, we trained two feed-forward models: one with 4 hidden layers of 256 units each and other with 11 hidden layers with 128 units each. The idea of training two models is to evaluate the network depth and ensembling in final performance. This architecture were found by some random search regarding of hidden layers and units, evaluating accordingly to an online metric we designed to take in account accuracy and target error.

\section{Results and Discussion}\label{sec:results_and_discussion}

In this work, we address the problem of humanoid robot kick motion learning for experimentation. We describe the task in Appendix \ref{ap:kick}. In order to evaluate the BUMPS algorithm in terms of efficiency to achieve passing-level control, we need to answer three questions:

\begin{itemize}
	\item \textbf{How BUMPS algorithm is compared with model-free RL to learn all tasks? And for adapting to new ones?} -- This question validates the initial idea of considering precise kick as a meta-learning problem;
	\item \textbf{How much meta-training is able to generalize to unseen tasks?} -- This question validates our assumptions and hypotheses described in Section \ref{sec:bumps};  and
	\item \textbf{How much policy filtering improves meta-policy generalization?} -- This question validates the idea of contextual gradients and, jointly to second question, validates our explanation from Section \ref{sec:bumps}.

\end{itemize}

To start answering the first question, we performed an RL training using PPO where, for each episode, we randomly sampled a task and then conduct the kick for this desired distance. The reward was the same as for the case of single task expert policy, considering the sampled distance. We also used the same hyperparameters except for the network architecture, where we used the same model for meta-policy with 11 hidden layers with 256 units each. We firstly imitated a 12m kick, to ensure fair initialization. 

\begin{figure}[!htpb]
	\centering
	\includegraphics[ width=0.68\textwidth]{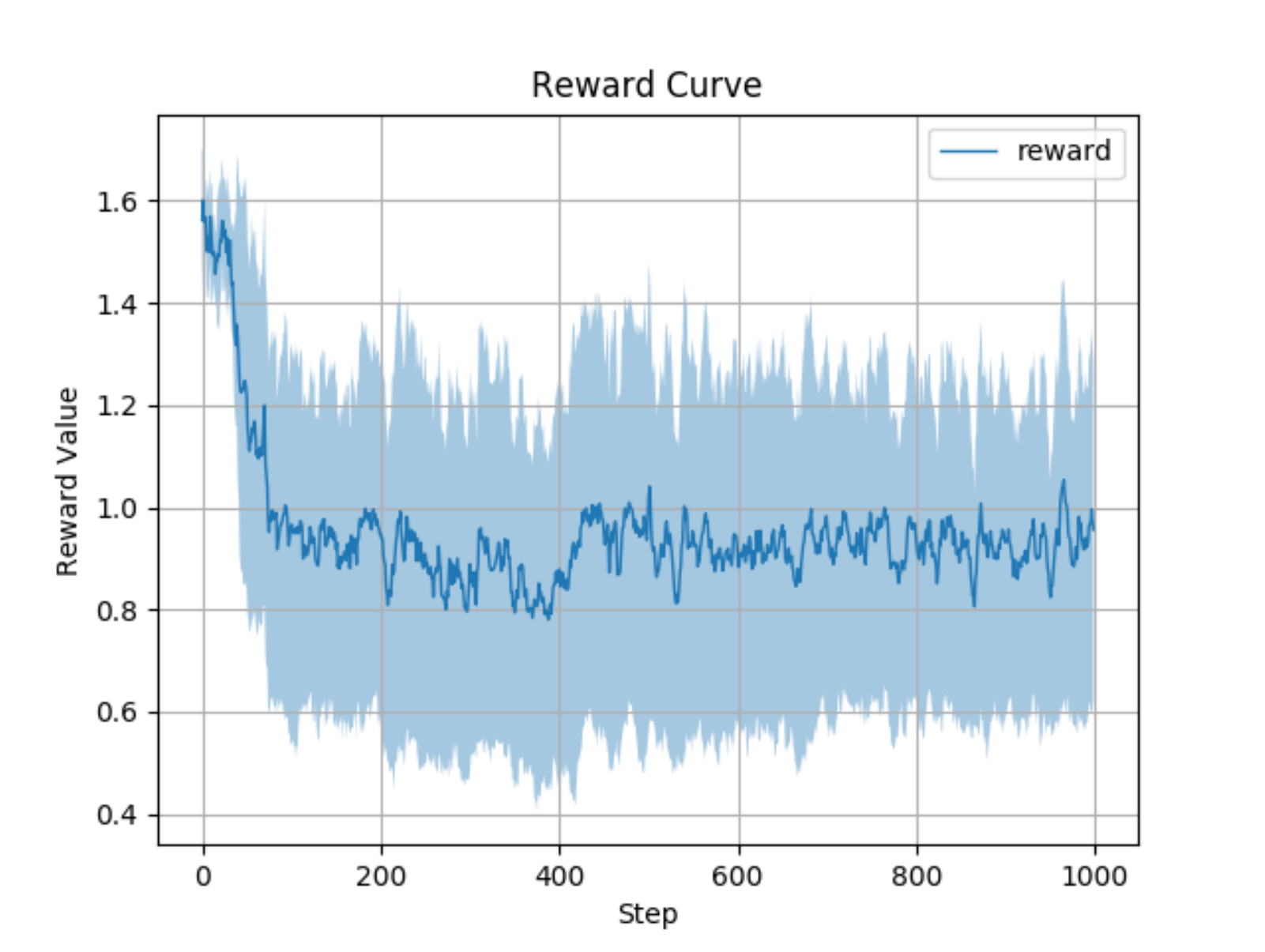}
	\caption{Reward curve (with 95\% bootstrapped confidence interval) during direct RL training of meta-policy.}
	\label{fig:rwcurvemetapolicyrl}
\end{figure}

We ran six experiments with different seeds and presented the reward curve in Figure \ref{fig:rwcurvemetapolicyrl}. As we can observe, in the context of our experiments, the RL training performed poorly, being not able to improve reward, but actually being catastrophic. On the other side, as we will see later, BUMPS is able to learn all meta-testing tasks.

In terms of second question, we can confront accuracy and mean error from single-task expert policies and meta-policy, using the evaluation scenario: one hundred repetitions of each kick policy. Table \ref{tab:comparisonmetapolicy} shows such values. In this experiment, we evaluate the single-task expert policies in their respective meta-training tasks, while the meta-policy in all meta-testing tasks. For meta-policy, we considered two networks with similar number of parameters: one with 4 hidden layers of 256 neurons and the other with 11 hidden layers with 128 neurons.

\begin{table}[!htbp]
	\caption{Meta-Policy Generalization Performance}
	\begin{center} 
		\begin{tabular}{|c|c|c|c|c|}
			\hline
			\textbf{Model}&\multicolumn{4}{|c|}{\textbf{Statistics}} \\
			\cline{2-5} 
			\textbf{Type} &  \multicolumn{2}{|c|}{\textbf{Accuracy}}& 
			\multicolumn{2}{|c|}{\textbf{Error (\(m\))}}\\
			\cline {2-5} 
			& \textbf{\textit{Mean}}& \textbf{\textit{Std}}
			& \textbf{\textit{Mean}}& \textbf{\textit{Std}} \\
			\hline
			Single-Task Expert Policies  & 0.89 & 0.04 & 0.57 & 0.14 \\
			\hline
			Meta-Policy (4, 256) & 0.84 & 0.10 & 0.72\footnote{hi} & 0.25  \\
			\hline
			Meta-Policy (11, 128) & 0.84 & 0.17 & 0.68 & 0.28\\
			\hline
		\end{tabular}
		
		\label{tab:comparisonmetapolicy}
	\end{center}

\end{table}
\footnotetext{In the case of meta-policies, for a fair comparison, we computed mean error considering only kicks with accuracy higher than 70\%, which means that there is a effective kick representation. For Single-Task Expert Policies, all kicks have accuracy higher than such threshold, thus we use all of them.}

As observed, the meta-policies are able to not only learn a kick representation for all meta-tasks, but also maintain a reasonable mean error and accuracy that are comparable to single-task expert policies. In fact, approximately 90\% of meta-policy kicks in both networks achieved more than 0.7 of accuracy (among them, approximately 90\% has error less than one meter).

Therefore, our experiments shows that a properly network can generalize to unseen tasks in the context of BUMPS algorithm. Nevertheless, this is not enough to solve all meta-testing tasks -- we can improve those values using Policy Filtering.

To answer the last question, we present the final results of the BUMPS algorithm, comparing with single-task expert policies. We present four models: 

\begin{itemize}
	\item \textbf{Filtered Meta-Policy (4, 256)} -- Final model where we sampled policy contexts from network with 4 hidden layers of 256 units after Policy Filtering;
	\item \textbf{Filtered Meta-Policy (11, 128)} -- Final model where we sampled policy contexts from network with 4 hidden layers of 128 units after Policy Filtering;
	\item \textbf{Filtered Ensemble} -- Final model where we sampled policy contexts from both previous meta-policy models;  and
	\item \textbf{Filtered Meta-Policy High Sample Rate (4, 256)} -- Final model where we sample policy contexts from network with 4 hidden layers of 256 units after Policy Filtering, but sampling 1091 tasks (approximately 10x more tasks than previous models).
\end{itemize}

Final results are presented in Table \ref{tab:finalmetapolicy}. As we observe, there is a visible improvement from meta-policy, especially when considering ensemble and a high sample rate. In all models, BUMPS achieved mean error less than a half meter, which is better than initial performance for meta-training tasks. Finally, we present a video to illustrate final policies for some tasks. \footnote{\url{https://youtu.be/RZs4GM7xObM}}

\begin{table}[!htbp]
	\caption{Final results for BUMPS algorithm}
	\begin{center} 
		\begin{tabular}{|c|c|c|c|c|}
			\hline
			 \textbf{Model}&\multicolumn{2}{|c|}{\textbf{Error (m)}}&\multicolumn{2}{|c|}{\textbf{Rel. Error (\%)}} \\
			\cline{2-5} 
			 \textbf{Type} & \textbf{Mean} & \textbf{Std} & \textbf{Mean} & \textbf{Std} \\
			\hline Policy
			 (4, 256)  & 0.45 & 0.10 & 3.8 & 1.1 \\
			\hline Policy
			  (11, 256) & 0.43 & 0.13 & 3.6 & 1.3 \\
			\hline
			 \textbf{Ensemble} & \textbf{0.40} & \textbf{0.11} & \textbf{3.3} & \textbf{1.1}  \\
			\hline
			\centering \textbf{High Sample Rate  (4, 256) }  & \textbf{0.40} & \textbf{0.09} & \textbf{3.3} & \textbf{0.9} \\
			\hline
		\end{tabular}
		
		\label{tab:finalmetapolicy}
	\end{center}
	
\end{table}

\section{Conclusion and Future Work}\label{sec:conclusion}

In this paper, we present a first version of BUMPS algorithm, which aims to reuse the knowledge from few single-task expert policies to learn a meta-policy for a distribution over such tasks and then improve sample complexity. Due to the lack of benchmark tasks with hard setup for meta-RL evaluation, we introduced the kick motion learning in RoboCup 3D Soccer simulation, showing that BUMPS is valuable to apply in the context of Robotics, by learning robust policies for a diversity of skills, from a variety of possible data sources. Specifically, we highlight some advantages of BUMPS:

\begin{itemize}
	\item \textbf{Simplicity} -- BUMPS relies on first-order derivatives to train a meta-policy in a standard way, by exploring the relationship between the geometric representation of task contextualization and optimal solution manifolds;
	\item \textbf{Sample Efficiency} -- In contrast to other methods for meta-policy search where meta-training also performs RL, BUMPS only needs to interact with environment during RL training of single tasks expert policies and contexts' evaluation; and
	\item \textbf{Model Agnostic} -- BUMPS is model agnostic, which means that it does not rely on a specific architecture to work. Nevertheless, this property can be exploited to improve generalization by exploiting task structure. We let as future work.
\end{itemize} 

As continuation of this work, we plan to address three ideas:

\begin{itemize}
    \item \textbf{Benchmark Evaluation}: We plan to evaluate BUMPS in other benchmark environments and tasks presented in the Literature, such as MuJoCo meta-RL tasks \cite{DBLP:journals/corr/FinnAL17}, in order to compare with other methods for meta-policy search;
    \item \textbf{Learning Task Embeddings}: We plan to address the geometric constraint needed from the context variable to exploit optimal solution manifolds, by learning task embeddings, with a proper network or optimization procedure; and
    \item \textbf{Bayesian Contextual Policies}: We plan to modify the meta-policy to consider each contextual policy as a random variable where we sample actions during meta-testing in order to filter a final policy to that specific task.
\end{itemize}

\section{Acknowledgments}

We would like to acknowledge Intel and the Student Ambassador for AI program for providing all the computational resources and necessary platform to execute this research through Intel AI DevCloud.

We also would like to acknowledge Deep Learning Brazil research group for all financial support and insightful discussions throughout this work. Finally, we are also grateful to ITA and all the ITAndroids team, especially 3D Soccer simulation team members for the hard work in the development of the base code.

\bibliographystyle{plainnat}
\bibliography{references}

\newpage
\begin{appendices}
\section{Task Description}\label{ap:kick}

\begin{figure}[!htbp]
	\centering
	\includegraphics[ width=0.8\textwidth]{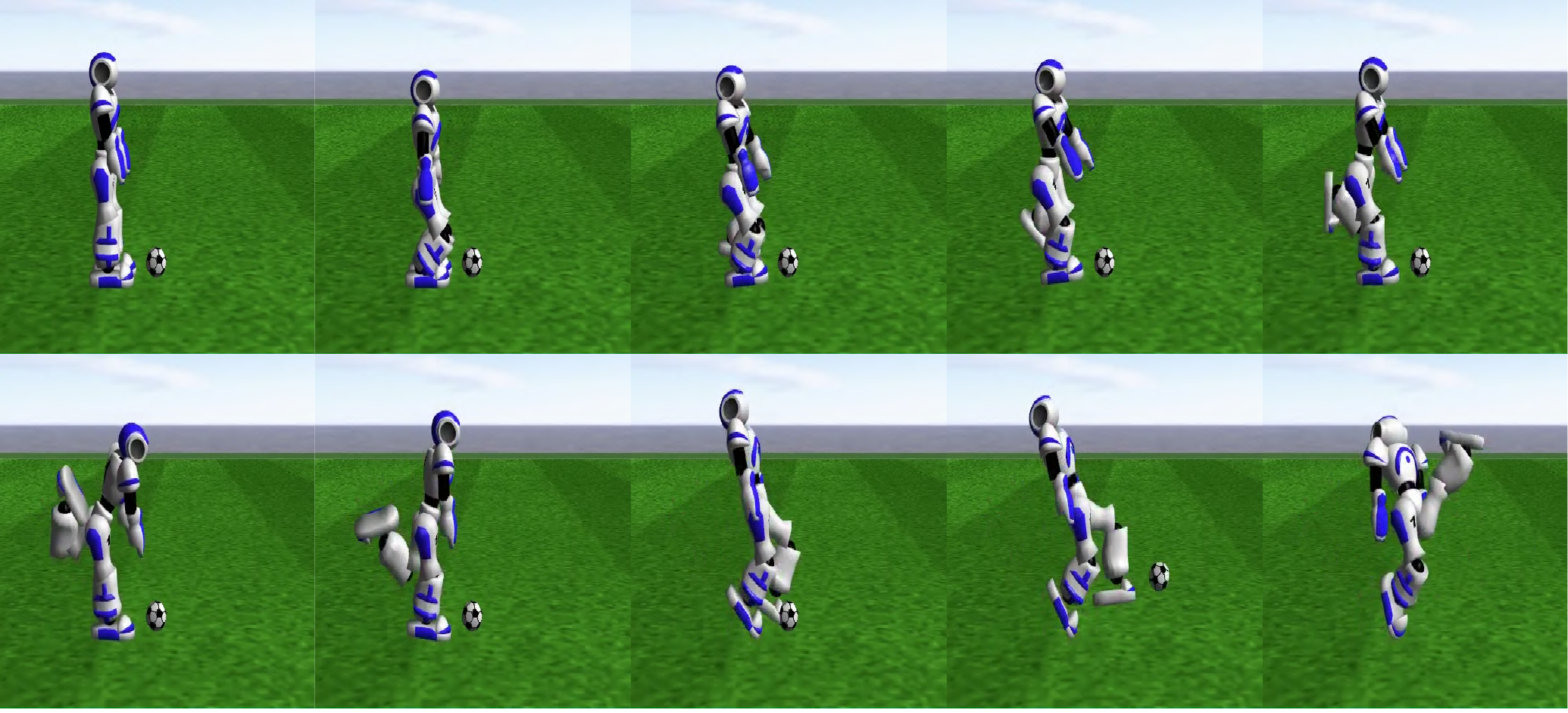}
	\caption{Illustration of Kick task. We consider the environment and task a hard setup, because we lead with a high dimensional continuous domain with 22 DoF and the stochasticity from the simulator. Furthermore, the reward is very delayed, since the agent is required to perform all the motion and hit the ball before receiving the reward, and all actions during this moment do not influence the final reward.}
	\label{fig:kick}
\end{figure}

We address the problem of humanoid robot kick motion learning to evaluate BUMPS algorithm, presented in Figure \ref{fig:kick}. Specifically, our objective is to learn how to learn to kick: given a range of possible distances for a kick (as a distribution of tasks), we can sample a policy that precisely kick for each task.

We use RoboCup 3D Soccer Simulation environment, a stochastic simulation environment based on SimSpark, a generic physical multi-agent system simulator. SimSpark uses the Open Dynamics Engine (ODE) library for its realistic simulation of rigid body dynamics with collision detection and friction. For further details, we refer to \cite{LNAI12-MacAlpine2}. In this environment, the agent is a simulated NAO robot.

In this task, the agent receives a counter variable that indicates the current time step and the target distance where the ball -- initially placed 0.2 meter in front of the agent -- should achieve after kick. The action space is a set of 22 joint desired positions. 

The distribution over tasks (i.e, kick target distances) is defined uniformly in the interval between 7 and 18 meters. The meta-training tasks are spaced by half a meter (i.e, 7.0, 7.5, $\dots$, 17.5, 18.0) and the meta-testing tasks spaced by 0.1 meter (7.1, 7.2, 7.3, $\dots$, 17.9, 18.0). The final evaluation was done by considering the final target error.

Finally, the reward to train single-task expert policies is given by a scaled absolute distance between the current ball position and the target, at each time step.

\newpage
\section{Hyperparameters}\label{ap:hypers}
For further details about the experiments, we refer to the available code.\footnote{\url{https://github.com/luckeciano/bumps}}

\subsection{Single-Task Expert Policies}

\begin{table}[!htbp]
\caption{Hyperparameters -- PPO}
	\begin{center} 
		\begin{tabular}{|c|c|c|c|c|c|}
			\hline
			Hyperparameter & Value   \\
			\hline
			Learning Rate &    $10^{-6}$    \\
			Timesteps per Actorbatch & 4096    \\
			Batch Size &   1024     \\
			Epochs &  30\\
			$\gamma$ & 0.999 \\
			$\lambda$ & 1.0 \\
			Timesteps (Total) & $12 \times 10^{6}$ \\
			Clip Parameter & 0.29\\
			Entropy Coefficient & 0.01 \\
			Reward Multiplicative Factor & 0.1 \\
			\hline
		\end{tabular}
		
		\label{tab:hypersprecisekick}
	\end{center}
\end{table}
\subsection{Meta-Policy Training}

\begin{table}[!htbp]
\caption{Hyperparameters -- Meta-Policy Training}
	\begin{center} 
		\begin{tabular}{|c|c|c|c|c|c|}
			\hline
			Hyperparameter & Value   \\
			\hline
			Learning Rate &    $3^{-6}$    \\
			Activations & $tanh$    \\
			Batch Size &   128     \\
			Epochs &  150000\\
			Learning Rate Decay & 0.8 \\
			\hline
		\end{tabular}
		
		\label{tab:hypersprecisekick}
	\end{center}
\end{table}

\end{appendices}
\end{document}